\documentclass[11pt]{article}

\usepackage[preprint]{acl}

\usepackage{times}
\usepackage{latexsym}

\usepackage[T1]{fontenc}
\usepackage[utf8]{inputenc}

\usepackage{microtype}

\usepackage{inconsolata}

\usepackage{graphicx}

\usepackage{tabularx}
\usepackage{pdflscape}
\usepackage{authblk}
\usepackage{etoolbox}
\usepackage{booktabs}
\usepackage{multirow}
\usepackage{makecell}
\usepackage{xcolor}

\definecolor{kh}{HTML}{168aff}

\newcommand{\myparagraph}[1]{\vspace{0.25em}\noindent\textbf{#1} \ }

\newcommand{\domainlist}{}

\newcommand{\defdomain}[2]{\csdef{dom#1}{#2}}
\defdomain{tamu}{@tamu.edu}
\defdomain{asml}{@asml.com}

\newcommand{\registeremail}[2]{%
  \ifinlist{#2}{\domainlist}{}{\listadd{\domainlist}{#2}}%
  \ifcsdef{list#2}{\csxappto{list#2}{, #1}}{\csdef{list#2}{#1}}%
}

\newcommand{\showemails}{%
  \renewcommand*{\do}[1]{\{\csuse{list##1}\}\csuse{dom##1}\quad}%
  \dolistloop{\domainlist}%
}

\makeatletter
\newcommand{\printcoauthornote}{%
  \begingroup
  \def\@thefnmark{*}%
  \@footnotetext{These authors contributed equally to this work.}%
  \endgroup
}
\makeatother
\title{ConstructCIE: A Dataset for Extracting Causal Information from Construction Accident Narratives}

\author[$\dagger$]{\textbf{Hung Nguyen}} \registeremail{nguye3hv}{tamu}
\author[$\ddagger$]{\textbf{Jaehoon Lee}} \registeremail{jlee1215}{tamu}
\author[$\ddagger$]{\textbf{Namgyun Kim}} \registeremail{nkim}{tamu}
\author[$\dagger$]{\textbf{Kuan-Hao Huang}} \registeremail{khhuang}{tamu}

\affil[$\dagger$]{Department of Computer Science, Texas A\&M University}
\affil[$\ddagger$]{Department of Construction Science, Texas A\&M University}

\newsavebox{\emailbox}
\sbox{\emailbox}{\texttt\showemails}

\affil[ ]{\usebox{\emailbox}}

\begin{document}
\maketitle
% \printcoauthornote % <-- THE FIX: Call this immediately after \maketitle
\begin{abstract}

Construction accident narratives contain rich causal information, but the evidence is often implicit, long-span, and distributed. We introduce \textsc{ConstructCIE}, a manually annotated dataset for Causal Information Extraction from OSHA construction accident reports. The dataset uses a hierarchical schema for accident types, causal factors, sub-causal factors, and supporting evidence spans. We evaluate supervised sequence taggers and instruction-tuned LLMs in an end-to-end hierarchical extraction setting. Results show that most evaluated models achieve strong accident-type prediction and recover broad causal meaning but remain limited in precise span-level extraction. Joint Hierarchical Extraction generally achieves stronger exact and soft matching, while Individual Hierarchical Extraction sometimes achieves higher keyword F1. Error distributions vary by extraction strategy, but evidence-selection and span-boundary errors remain common. These findings show that reliable Causal Information Extraction for construction accidents requires stronger domain grounding and more accurate evidence extraction. The code and data can be found at \url{https://github.com/lab-flair/ConstructCIE}.

\end{abstract}

\section{Introduction}

% Introduce about accidents
% Why accident extraction is difficult 

Construction is economically essential but remains highly hazardous. In 2024, construction accounted for 1,034 of 5,070 work-related fatalities in the United States, the highest among private industries \cite{bls2026cfoi_a, bls2026cfoi_b}. Although construction represented only about 5\% of U.S. nonfarm employment, it was responsible for roughly 20\% of worker deaths \cite{bls2026ces_c}. Accident reports are valuable for understanding causes and preventing future incidents \cite{rupasinghe2020, zou2017}. However, these narratives are difficult to analyze because they describe dynamic site conditions, changing work processes, and causal chains linking unsafe conditions, procedural failures, worker actions, and injury outcomes \cite{lee2012, zhou2014}. Thus, much safety knowledge remains unstructured and requires manual analysis \cite{zou2017, ma2024}.

%Why are existing datasets no good
% \kh{There is a gap here. We should discuss why this can be viewed as an event extraction. And what's the difference too.}

This task aligns with Event Extraction (EE) but differs from conventional trigger-centered EE. Existing datasets such as ACE05 \cite{ace} focus on explicit event triggers and predefined argument roles, whereas accident narratives require identifying why an accident occurred through long, sometimes implicit causal evidence distributed across multiple sentences \cite{hershowitz2024, chen2025tag}. Thus, existing EE datasets are insufficient for Causal Information Extraction (CIE) for construction accidents, which requires domain terminology, engineering context, and ambiguous causal language \cite{chen2024, norouzi2025, fu2026}.

\begin{figure}
    \centering
    \includegraphics[width=1\linewidth]{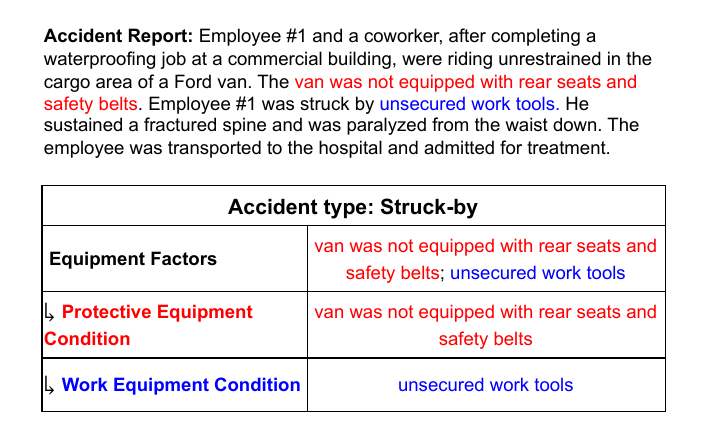}
    % \caption{An example of an accident report and extracted \textit{Equipment Factors} causal factor which contains \textit{Protective Equipment Condition} and \textit{Work Equipment Condition} sub-causal factors.}
    \caption{An example accident report and its extracted \textit{Equipment Factors}, including the \textit{Protective Equipment Condition} and \textit{Work Equipment Condition} sub-causal factors.}
    \label{fig:pred_example}
    % \vspace{-1em}
\end{figure}

% \kh{The first three paragraphs take to much space. Try to make them shorter.}

% \kh{We don't need to list all the dataset name.}
% \kh{no existing datasets are suitable for extracting causal chains from construction events ==> what it matters? We need to discuss.}

% Task Proposal and Dataset Proposal 
% Why is this task and dataset special

To address these limitations, we introduce \textsc{ConstructCIE}, a construction-specific dataset for CIE from accident narratives. Constructed from 530 Occupational Safety and Health Administration (OSHA) accident investigation summaries published between 2011 and 2023 \cite{osha}, \textsc{ConstructCIE} supports accident-type prediction and hierarchical extraction of causal factors, sub-causal factors, and supporting evidence. Unlike conventional event extraction datasets, it targets triggerless, long-span, and potentially discontinuous causal evidence in construction accident analysis.

This work makes three main contributions: (1) We introduce \textsc{ConstructCIE}, a construction-specific CIE dataset built from OSHA accident narratives with annotations for accident types, causal factors, sub-causal factors, and supporting evidence. (2) We propose a hierarchical construction accident causal schema covering accident types, broad causal factors, and fine-grained sub-causal factors. (3) We evaluate supervised and LLM-based methods end-to-end under joint and individual hierarchical extraction strategies, revealing a persistent gap between broad causal recovery and precise evidence extraction.

% \kh{If our conclusion is current models are good, it will largely reduce the necessity of constructing this dataset. A better claim is models are still not good enough for end2end? And there is a gap between supervised baseline (is it true?), suggesting additional effort needed for making LLMs stronger in domain knowledge.}
% \kh{The current intro is too long and too detailed.}

\section{Related Work}

\myparagraph{Construction Accident Ontologies}
Early works in safety science defined rich theoretical taxonomies for describing the complex and interactive factors underlying systemic accident causation \cite{perrow1984, reason1997}. Subsequent analytical frameworks, such as HFACS \cite{hfacs} and STAMP \cite{stamp}, formalized hierarchical categories of human, organizational, and system-level risk factors to support accident analysis and communication among safety experts. Recent natural language processing studies have focused on organizing specific construction-related hazards, such as the Risk Knowledge Graph in Railway Safety (RKGRS) for railway accidents \cite{rkgrs}, as well as targeted entity dictionaries for struck-by accidents \cite{zhou2025}. However, existing event-extraction benchmarks such as ACE05 and WikiEvents rely on general-domain ontologies \cite{ace,wikievents}, while recent safety-domain extraction pipelines remain focused on localized, textually stated causal spans rather than dense multi-hop causal graph construction \cite{chen2026}. As a result, existing ontologies are often too theoretical, sentence-bound, or narrowly tailored to specific accident types, limiting their applicability to document-level Causal Information Extraction across diverse construction accidents.

\myparagraph{Causal Information Extraction}
Early work in CIE relied on pattern-matching and rule-based methods that used explicit linguistic cues to identify causal relations \cite{khoo1998, khoo2000}. With the introduction of transformer-based models, pre-trained language models such as BERT enabled more flexible modeling of complex and implicit causal relations \cite{bert}. A common approach is to encode the context containing two target event mentions with BERT and apply a classifier over the resulting event representations to predict whether the events are causally related \cite{liu2023}. However, simple event-pair context encoding may be insufficient, as it does not explicitly incorporate broader semantic structures, such as event-centric representations and event-associated paths, that help capture implicit causal dependencies between events \cite{hu2023}. Furthermore, real-world causal structures are often intertwined in complex many-to-one or one-to-many relationships \cite{chen2025}, and the arguments associated with causal events are often implicit or distributed across fragmented textual evidence rather than expressed as clearly bounded local spans \cite{discourseee}. To capture these complex interdependencies, recent studies have leveraged large language models (LLMs), automatic prompt engineering, and graph neural networks (GNNs) for multi-hop causal reasoning \cite{chen2025, yan2025}. Despite these advances, CIE remains under-explored in the construction safety domain. Prior construction safety research has predominantly relied on structured or tabular accident-report fields, while narrative accident reports remain comparatively underutilized despite containing richer contextual information about accident circumstances, root causes, and contributing factors \cite{unmesa2025}. This gap motivates the need for a construction-specific CIE benchmark that supports document-level extraction of causal factors and supporting evidence from unstructured accident narratives, especially when such information is implicit, fragmented, or distributed across multiple parts of the report.

% Early CIE methods relied on pattern matching and rules based on explicit causal cues \cite{khoo1998, khoo2000}. Transformer models such as BERT later enabled more flexible modeling of implicit causal relations \cite{bert}, often by encoding contexts containing two event mentions and classifying whether they are causally related \cite{liu2023}. However, event-pair encoding may overlook broader semantic structures, such as event-centric representations and event-associated paths, that support implicit causal reasoning \cite{hu2023}. Real-world causality also involves complex many-to-one and one-to-many relations \cite{chen2025}, with evidence often fragmented or distributed rather than expressed in clearly bounded local spans \cite{discourseee}. Recent work therefore uses LLMs, prompt engineering, and GNNs for multi-hop causal reasoning \cite{chen2025, yan2025}. Yet CIE remains underexplored in construction safety, where prior work relies mainly on structured fields despite the richer contextual information in narrative reports \cite{unmesa2025}. This motivates a construction-specific benchmark for document-level extraction of causal factors and supporting evidence from unstructured accident narratives.

\begin{table*}[t]
\centering
\scriptsize
\setlength{\tabcolsep}{2pt}
\renewcommand{\arraystretch}{0.95}
\begin{tabularx}{\textwidth}{
>{\raggedright\arraybackslash}p{0.12\textwidth}
>{\raggedright\arraybackslash}p{0.13\textwidth}
>{\raggedright\arraybackslash}X
>{\centering\arraybackslash}p{0.13\textwidth}}
\toprule
\textbf{Causal Factor} 
& \textbf{Sub-causal Factor} 
& \textbf{Definition}
& \textbf{\shortstack{Sub-causal\\Factor Type}} \\
\midrule

Working Circumstances
& Construction Trade
& Working Circumstances describe the physical and operational context of the accident, while Construction Trade identifies the primary construction activity being performed.
& Classification \\

\midrule

Object Involved
& N/A
& Object Involved captures the equipment, structure, material, or object that physically interacted with or contributed to the accident mechanism.
& N/A \\

\midrule

\multirow{2}{=}{Managerial Factors}
& Failure of Hazard Management
& Managerial Factors capture organizational or supervisory deficiencies, while this subfactor identifies failures to recognize foreseeable hazards during planning or risk assessment.
& Extraction \\
\cmidrule(lr){2-4}
& Deficiency in Safety Training
& Managerial Factors capture organizational or supervisory deficiencies, while this subfactor identifies inadequate task-specific safety training.
& Extraction \\

\midrule

\multirow{2}{=}{Equipment Factors}
& Protective Equipment Condition
& Equipment Factors capture equipment-related contributors, while this subfactor identifies missing, unavailable, inadequate, or improperly configured protective equipment.
& Extraction \\
\cmidrule(lr){2-4}
& Work Equipment Condition
& Equipment Factors capture equipment-related contributors, while this subfactor identifies unsafe conditions, defects, configuration problems, or availability issues involving tools or machinery.
& Extraction \\

\midrule

\multirow{2}{=}{Working Condition Factors}
& Weather Condition
& Working Condition Factors capture physical work-environment contributors, while this subfactor identifies weather conditions affecting the work environment or accident sequence.
& Extraction \\
\cmidrule(lr){2-4}
& Workspace Condition
& Working Condition Factors capture physical work-environment contributors, while this subfactor identifies unsafe spatial, structural, surface, support, or layout conditions of the workspace.
& Extraction \\

\midrule

\multirow{2}{=}{Behavioral Factors}
& Inattentive Behavior
& Behavioral Factors capture task-level human actions or lapses, while this subfactor identifies distraction, reduced vigilance, or loss of situational awareness.
& Extraction \\
\cmidrule(lr){2-4}
& Noncompliant Behavior
& Behavioral Factors capture task-level human actions or lapses, while this subfactor identifies disregard of safety rules, safe work practices, or required protective measures.
& Extraction \\

\midrule

\multirow{2}{=}{Consequences}
& Severity
& Consequences capture accident outcomes, while this subfactor classifies the seriousness of injury from minor injury to fatality.
& Classification \\
\cmidrule(lr){2-4}
& Affected Body Part
& Consequences capture accident outcomes, while this subfactor identifies the body part or parts harmed in the incident.
& Extraction \\

\bottomrule
\end{tabularx}
\caption{Causal factor taxonomy, row-level definitions, and sub-causal factor types for construction accidents. Detailed definitions are provided in Appendix~\ref{sec:extraction_framework}.}
\label{tab:causal_taxonomy}
% \vspace{-1em}
\end{table*}
% \kh{[KH: If no sub factor, we should put N/A]} 
% \kh{[KH: All the subfactor is extraction? We can put the type of task as one additional column indicating extraction or classification.]}

\section{\textsc{ConstructCIE} Curation}
% \kh{I suggest using title ``'ConstructCIE'}
\subsection{Task Definition}

% An \textbf{accident type} represents the high-level accident mechanism. A \textbf{causal factor} is a condition, action, failure, or circumstance explaining how or why an accident occurred. A \textbf{sub-causal factor} is a fine-grained category specifying the type of causal evidence. Each report is also associated with an \textbf{accident type}, which represents the high-level accident category, such as fall, struck-by, caught-in/between, or electrocution.

An \textbf{accident type} represents the high-level accident mechanism. We consider four accident types: \textit{fall}, \textit{struck-by}, \textit{caught-in/between}, and \textit{electrocution}. A \textbf{causal factor} is a condition, action, failure, or circumstance explaining how or why an accident occurred. A \textbf{sub-causal factor} is a fine-grained category specifying the type of causal evidence.

% \kh{Do we need to predict accident types? It's not introduced.}

Given an accident narrative, CIE aims to produce a hierarchy of accident type, causal factors, sub-causal factors, and evidence. In the end-to-end setting, the model first predicts the accident type from the report. It then identifies the causal factors present in the narrative and extracts their supporting text span(s). For each causal factor, the model further predicts the associated sub-causal factor information. Depending on the sub-causal factor, this step is formulated either as classification or span extraction. For example, in Figure~\ref{fig:pred_example}, the causal factor \textit{Equipment Factors} is supported by evidence that can be further divided into \textit{Protective Equipment Condition} and \textit{Work Equipment Condition}.

% \kh{Add a paragraph discussing what's the challenge, and the difference from EE. This should be detailed, and the version we mention in intro should be high-level.}

This task differs from conventional trigger-centered event extraction. In standard EE, models usually identify explicit event triggers and extract local arguments associated with those triggers \cite{ace}. In contrast, in construction accident narratives, the key information is not only what happened, but why it happened, with causal evidence being expressed through long spans distributed across multiple sentences \cite{hershowitz2024, chen2025tag}.  Moreover, a single causal factor may be supported by evidence that must be further decomposed into multiple sub-causal factors, and some sub-causal factors require classification rather than direct span extraction. This requires connecting accident context, domain terminology, and implicit causality.

% \kh{I think we need a figure show one example, and all the terms we define.}

\subsection{Causal Schema Creation}
% \kh{We should not call it event.}
% \kh{The current writing is not clear to me what should be predicted. Should we classify causal factor or they are given?}
% \kh{What's the task for object Involved?}
% \kh{You said six, but there are seven in table.}

% We define CIE using a construction-specific causal schema specifying what to extract from each narrative. The schema has two levels: \textit{causal factors}, which represent broad accident contributors, and \textit{sub-causal factors}, which capture fine-grained conditions, failures, behaviors, or outcomes. 
%Short names for both levels are used in the result tables for compact reporting.

We define CIE using a construction-specific hierarchical framework. Each report is assigned a top-level accident type (\textbf{Acc.}), followed by a two-level causal schema. The accident type represents the overall accident mechanism, causal factors represent broad accident contributors, and sub-causal factors capture fine-grained conditions, failures, behaviors, or outcomes.

As shown in Table~\ref{tab:causal_taxonomy}, the schema contains seven primary causal factors: \textit{Working Circumstances}, \textit{Object Involved}, \textit{Managerial Factors}, \textit{Working Condition Factors}, \textit{Equipment Factors}, \textit{Behavioral Factors}, and \textit{Consequences}. Their short names are reported as \textbf{Work. Circ.}, \textbf{Obj. Inv.}, \textbf{Mgmt.}, \textbf{Cond.}, \textbf{Equip.}, \textbf{Behav.}, and \textbf{Conseq.}, respectively. The schema further defines sub-causal factors, such as \textit{Construction Trade} (\textbf{Constr. Trade}), \textit{Workspace Condition} (\textbf{Workspace Cond.}), \textit{Protective Equipment Condition} (\textbf{Prot. Equip.}), \textit{Work Equipment Condition} (\textbf{Work Equip.}), \textit{Inattentive Behavior} (\textbf{Inatt. Behav.}), \textit{Noncompliant Behavior} (\textbf{Noncomp. Behav.}), and \textit{Affected Body Part} (\textbf{Body Part}). Table~\ref{tab:causal_taxonomy} provides all factor definitions and prediction types.

The schema is drawn from construction safety literature on managerial, environmental, equipment-related, and behavioral accident contributors \cite{luo2021,shan2023,kang2018,halperin2004141,jahangiri201988,wong2009}. Given an accident narrative, a model identifies causal factors and extracts or classifies sub-causal information using this schema. 
% This supports high-level causation analysis and fine-grained evidence extraction.

% \section{Data Curation and Analysis}

% \kh{This section should be merged to the previous one.}

\subsection{Data Curation}

We constructed \textsc{ConstructCIE} from OSHA accident investigation summaries published between 2011 and 2023 \cite{osha}. We then applied a two-stage screening process: record-level screening removed duplicate records and reports unrelated to construction activities, while content-level screening retained only reports whose narratives explicitly described both accident circumstances and causes. This yielded 530 reports.

% The resulting reports were manually annotated by two domain experts in construction safety using the taxonomy shown in Table~\ref{tab:causal_taxonomy}. For each report, annotators labeled the accident type, causal factors, sub-causal factors, and supporting text spans. Annotations were cross-checked, and disagreements were resolved by consensus. Finally, all date information was removed using regular expression pattern matching to prevent models from learning false temporal patterns or hallucinating causal links based on timestamps.

The resulting 530 reports were independently annotated by two domain experts in construction safety using the taxonomy shown in Table~\ref{tab:causal_taxonomy}. For each report, both annotators labeled the accident type, causal factors, sub-causal factors, and supporting text spans. Disagreements were resolved through discussion until consensus. Finally, all date information was removed using regular-expression pattern matching to prevent models from learning false temporal patterns or hallucinating causal links based on timestamps.

To quantify annotation reliability before consensus reconciliation, we computed macro-averaged pairwise F1 for extraction and Cohen's kappa for classification across all 530 independently double-annotated reports. Pairwise F1 was 0.9213 for causal factors, 0.9562 for sub-causal factors, and 0.9410 overall, while Cohen's kappa was 1.0000 for accident type, 0.9686 for sub-causal factors, and 0.9784 overall. 
% These scores indicate a high level of agreement between the two experts before reconciliation.

% To construct this dataset, we first collected OSHA accident investigation summaries published between 2011 and 2023 \cite{osha}. Then, the dataset was filtered through a two layer-process. We first used (1) record-level screening, where  duplicated records and records not related to construction activities were removed. Afterwards, (2) content-level screening was used to only retain reports whose narratives explicitly described both the circumstances and the causes of the accident. This resulted in 530 accident reports. To identify the causal factors and sub-causal factors associated with each report, two domain experts in construction safety independently labeled each report using the taxonomy shown in Table~\ref{tab:causal_taxonomy}. The annotated labels are cross-checked, with disagreements being resolved through discussion until a consensus was reached. Finally, to prevent the any models from establishing false temporal biases or hallucinating causal links based on timestamps, all date information was explicitly removed from the text using regular expression pattern matching.
% \kh{The current writing does not emphasize it's manually annotated.}

\begin{table}[tbp]
    \centering
    \resizebox{.7\columnwidth}{!}{
    \begin{tabular}{lc}
        \toprule
        \textbf{Accident Type} & \textbf{Percentage (\%)} \\
        \midrule
        Fall & 40.6 \\
        Caught-in/between & 22.3 \\
        Struck-by & 22.1 \\
        Electrocution & 15.1 \\
        \bottomrule
    \end{tabular}}
    \caption{Accident Type Distribution}
    \label{tab:event_pct}
    % \vspace{-1em}
\end{table}

\begin{table}[tbp]
    \centering
    \resizebox{0.85\columnwidth}{!}{
    \begin{tabular}{lc}
        \toprule
        \textbf{Causal Factor} & \textbf{Percentage (\%)} \\
        \midrule
        Working Circumstances & 100 \\
        Object Involved & 94.5 \\
        Managerial Factors & 6.4 \\
        Working Condition Factors & 40.2 \\
        Equipment Factors & 26.6 \\
        Behavioral Factors & 41.5 \\
        Consequences & 99.2 \\
        \bottomrule
    \end{tabular}}
    \caption{Causal Factor Distribution}
    \label{tab:factor_comp}
\end{table}

\begin{table}[t]
    \centering
    \scriptsize
    \setlength{\tabcolsep}{3pt}
    \resizebox{0.87\columnwidth}{!}{%
    \begin{tabular}{lc}
        \toprule
        \textbf{Sub-causal Factor} & \textbf{Percentage (\%)} \\
        \midrule
        Construction Trade & 98.9 \\
        Failure of Hazard Management & 3.8 \\
        Deficiency in Safety Training & 2.8 \\
        Weather Condition & 7.0 \\
        Workspace Condition & 34.3 \\
        Protective Equipment Condition & 5.1 \\
        Work Equipment Condition & 22.1 \\
        Inattentive Behavior & 13.0 \\
        Noncompliant Behavior & 28.7 \\
        Severity & 99.2 \\
        Affected Body Part & 65.5 \\
        \bottomrule
    \end{tabular}%
    }
    \caption{Sub-causal Factor Distribution}
    \label{tab:subfactor_comp}
    % \vspace{-1em}
\end{table}

\subsection{Data Statistics}
% \kh{I suggest using Data Statistics.}
We analyze the distribution of accident types, causal factors, and sub-causal factors. Table~\ref{tab:event_pct} shows that \textit{Fall} accidents are the most common, followed by \textit{Caught-in/between} and \textit{Struck-by}; \textit{Electrocution} is least common.

% From Table~\ref{tab:factor_comp}, the dataset contains high coverage for \textit{Working Circumstances}, \textit{Object Involved} and \textit{Consequences}, with 98.9\% of reports containing working circumstance information, 94.5\% containing \textit{Object Involved} and 82.7\% containing \textit{Consequences}. Among causal factor categories, \textit{Behavioral Factors} and \textit{Working Condition Factors} appear most frequently, covering 20.8\% and 20.5\% of reports, respectively. \textit{Equipment Factors} appear in 13.8\% of reports, while \textit{Managerial Factors} appear much less frequently, covering only 3.3\% of reports. Thus, narratives more often describe site conditions, behaviors, and outcomes than managerial causes.

% From Table~\ref{tab:factor_comp}, the dataset contains high coverage for \textit{Working Circumstances}, \textit{Object Involved}, and \textit{Consequences}, which appear in 98.9\%, 94.5\%, and 82.7\% of reports, respectively. Among the remaining causal categories, \textit{Behavioral Factors} and \textit{Working Condition Factors} appear most frequently, covering 20.8\% and 20.7\% of reports. \textit{Equipment Factors} appear in 13.6\% of reports, while \textit{Managerial Factors} are much less frequent, appearing in only 3.3\%. Thus, the narratives more often describe site conditions, behaviors, and outcomes than managerial causes.

From Table~\ref{tab:factor_comp}, the dataset contains high coverage for \textit{Working Circumstances}, \textit{Consequences}, and \textit{Object Involved}, which appear in 100.0\%, 99.2\%, and 94.5\% of reports, respectively. Among the remaining causal categories, \textit{Behavioral Factors} and \textit{Working Condition Factors} appear most frequently, covering 41.5\% and 40.2\% of reports. \textit{Equipment Factors} appear in 26.6\% of reports, while \textit{Managerial Factors} are much less frequent, appearing in only 6.4\%. Thus, the narratives more often describe accident circumstances, outcomes, site conditions, and behaviors than managerial causes.

% \kh{I feel that we should use table for fig 3 and fig 4.} 
% From Table~\ref{tab:subfactor_comp}, \textit{Construction Trade} is highly complete, appearing in 98.9\% of reports. \textit{Severity} is available for all reports, while \textit{Affected Body Part} appears in 65.5\% of reports. For causal subcategories, \textit{Workspace Condition} is the most frequent sub-causal factor, appearing in 34\% of reports, followed by \textit{Noncompliant Behavior} and \textit{Work Equipment Condition}. In contrast, \textit{Failure of Hazard Management}, \textit{Deficiency in Safety Training}, \textit{Weather Condition}, and \textit{Protective Equipment Condition} are less frequent, indicating stronger class imbalance for these categories.

% From Table~\ref{tab:subfactor_comp}, \textit{Construction Trade} appears in 98.9\% of reports, while \textit{Severity} is available for all reports and \textit{Affected Body Part} appears in 65.5\%. Among the causal subcategories, \textit{Workspace Condition} is the most frequent at 34.3\%, followed by \textit{Noncompliant Behavior} at 28.7\% and \textit{Work Equipment Condition} at 22.1\%. In contrast, \textit{Failure of Hazard Management}, \textit{Deficiency in Safety Training}, \textit{Weather Condition}, and \textit{Protective Equipment Condition} appear infrequently, indicating substantial class imbalance for these categories.

From Table~\ref{tab:subfactor_comp}, \textit{Severity} and \textit{Construction Trade} appear in 99.2\% and 98.9\% of reports, respectively, while \textit{Affected Body Part} appears in 65.5\%. Among the causal subcategories, \textit{Workspace Condition} is the most frequent at 34.3\%, followed by \textit{Noncompliant Behavior} at 28.7\% and \textit{Work Equipment Condition} at 22.1\%. In contrast, \textit{Failure of Hazard Management}, \textit{Deficiency in Safety Training}, \textit{Weather Condition}, and \textit{Protective Equipment Condition} appear infrequently, indicating substantial class imbalance for these categories.

\begin{table*}[t]
\centering
\scriptsize
\setlength{\tabcolsep}{2pt}
\renewcommand{\arraystretch}{0.95}
\resizebox{\textwidth}{!}{%
\begin{tabular}{l c c ccc ccc ccc ccc ccc ccc ccc}
\toprule
\multirow{2}{*}{\textbf{Model}}
& \multirow{2}{*}{\textbf{$k$}}
& \multirow{2}{*}{\textbf{Acc.}}
& \multicolumn{3}{c}{\textbf{Work. Circ.}}
& \multicolumn{3}{c}{\textbf{Mgmt.}}
& \multicolumn{3}{c}{\textbf{Cond.}}
& \multicolumn{3}{c}{\textbf{Equip.}}
& \multicolumn{3}{c}{\textbf{Behav.}}
& \multicolumn{3}{c}{\textbf{Conseq.}}
& \multicolumn{3}{c}{\textbf{Obj. Inv.}}
\\
\cmidrule(lr){4-6}
\cmidrule(lr){7-9}
\cmidrule(lr){10-12}
\cmidrule(lr){13-15}
\cmidrule(lr){16-18}
\cmidrule(lr){19-21}
\cmidrule(lr){22-24}
& & 
& E & S & K
& E & S & K
& E & S & K
& E & S & K
& E & S & K
& E & S & K
& E & S & K \\
\midrule
TagPrime-C & - & 94.7 & 61.3 & 79.8 & 79.0 & \textbf{86.4} & \textbf{86.4} & \textbf{93.0} & 37.6 & 39.8 & 44.3 & \textbf{49.4} & \textbf{49.4} & 61.8 & \textbf{50.7} & 55.9 & 70.3 & 78.5 & 83.9 & 89.9 & 40.1 & 41.8 & 44.5 \\
TagPrime-CR & - & 93.6 & \textbf{63.4} & \textbf{84.2} & 78.1 & 66.3 & 72.0 & 79.8 & 34.8 & 36.2 & 38.3 & 38.9 & 41.9 & 56.7 & 50.3 & 53.6 & 69.4 & \textbf{79.3} & \textbf{86.7} & 89.9 & 7.3 & 7.3 & 11.5 \\
\midrule
Llama3.2-3B & 0 & 8.7 & 0.0 & 0.6 & 0.6 & 3.6 & 3.6 & 3.6 & 0.0 & 0.0 & 0.0 & 0.0 & 0.0 & 0.0 & 0.0 & 0.0 & 0.0 & 1.3 & 3.0 & 7.2 & 2.1 & 2.1 & 4.8 \\
Llama3.2-3B & 5 & 61.5 & 11.0 & 31.1 & 23.0 & 11.8 & 33.6 & 38.1 & 14.8 & 26.6 & 34.9 & 0.9 & 2.8 & 8.0 & 13.7 & 34.3 & 54.9 & 18.4 & 50.8 & 73.5 & 24.9 & 26.1 & 42.3 \\
Llama3.2-3B & 30 & 64.9 & 6.3 & 44.9 & 58.3 & 3.1 & 29.8 & 41.5 & 21.2 & 26.4 & 40.0 & 4.7 & 5.6 & 12.0 & 10.4 & 32.6 & 58.9 & 20.6 & 55.5 & 78.6 & 35.5 & 38.3 & 50.9 \\
\midrule
Llama3.2-11B & 0 & 83.8 & 0.4 & 4.3 & 2.7 & 0.0 & 23.1 & 23.1 & 5.6 & 10.5 & 14.9 & 2.5 & 6.2 & 14.8 & 4.3 & 26.4 & 51.9 & 5.4 & 11.5 & 32.5 & 10.3 & 17.4 & 39.7 \\
Llama3.2-11B & 5 & 81.1 & 31.4 & 60.6 & 79.5 & 28.4 & 59.7 & 76.3 & 20.4 & 23.3 & 32.0 & 12.3 & 14.3 & 25.7 & 20.2 & 40.6 & 76.9 & 27.5 & 50.0 & 84.6 & 45.6 & 47.1 & 59.4 \\
Llama3.2-11B & 30 & 81.1 & 28.3 & 60.8 & 83.8 & 16.9 & 35.4 & 57.1 & 20.8 & 25.6 & 29.1 & 14.2 & 18.3 & 34.0 & 23.4 & 43.7 & 73.2 & 35.9 & 58.4 & 88.8 & 54.1 & 56.0 & 66.1 \\
\midrule
Llama3.1-70B & 0 & 95.8 & 0.0 & 3.8 & 4.2 & 0.0 & 28.4 & 45.7 & 3.1 & 5.4 & 12.8 & 19.6 & 44.0 & 52.6 & 4.8 & 34.8 & 59.8 & 6.5 & 13.8 & 44.0 & 13.2 & 13.6 & 30.1 \\
Llama3.1-70B & 5 & 97.0 & 54.3 & 77.4 & 74.3 & 14.5 & 31.7 & 75.7 & 29.1 & 41.6 & 42.2 & 26.3 & 38.4 & 54.7 & 33.6 & \textbf{59.5} & 79.2 & 27.5 & 50.4 & 90.3 & 37.6 & 40.4 & 58.8 \\
Llama3.1-70B & 30 & 95.8 & 55.8 & 76.6 & 83.4 & 39.2 & 49.7 & 68.4 & \textbf{42.0} & \textbf{51.6} & 55.8 & 31.9 & 35.7 & 47.6 & 32.8 & 59.5 & \textbf{81.4} & 34.1 & 54.3 & 90.9 & 56.7 & \textbf{59.4} & 74.1 \\
\midrule
Qwen3.5-9B & 0 & 95.8 & 0.4 & 4.9 & 7.5 & 35.9 & 51.1 & 56.8 & 10.8 & 16.7 & 26.3 & 10.4 & 15.7 & 21.1 & 14.6 & 46.0 & 76.0 & 14.1 & 26.7 & 56.8 & 16.2 & 17.4 & 43.3 \\
Qwen3.5-9B & 5 & 92.8 & 37.4 & 70.2 & 92.8 & 19.7 & 39.4 & 79.9 & 19.6 & 31.6 & 44.2 & 17.9 & 25.7 & 44.4 & 33.2 & 55.9 & 74.2 & 22.6 & 55.9 & 91.7 & 41.4 & 44.5 & 66.1 \\
Qwen3.5-9B & 30 & 94.0 & 46.0 & 74.3 & \textbf{94.0} & 17.2 & 50.7 & 77.2 & 21.0 & 39.7 & 51.7 & 21.5 & 34.2 & 47.3 & 37.8 & 54.9 & 76.3 & 38.9 & 60.3 & \textbf{94.6} & \textbf{56.8} & 59.1 & \textbf{75.9} \\
\midrule
Qwen3.5-27B & 0 & 97.4 & 0.0 & 2.3 & 3.0 & 46.7 & 55.6 & 70.4 & 14.0 & 20.5 & 29.0 & 32.6 & 37.6 & 71.3 & 9.9 & 33.5 & 54.8 & 10.7 & 26.7 & 59.8 & 17.8 & 17.8 & 32.0 \\
Qwen3.5-27B & 5 & \textbf{98.1} & 50.9 & 77.0 & 77.0 & 32.6 & 62.8 & 80.7 & 26.8 & 39.7 & 47.5 & 34.3 & 47.0 & \textbf{73.4} & 31.9 & 54.6 & 71.5 & 33.6 & 59.5 & 87.3 & 32.5 & 36.0 & 54.9 \\
Qwen3.5-27B & 30 & 97.7 & 53.6 & 80.4 & 88.3 & 45.6 & 61.7 & 75.7 & 36.4 & 49.2 & \textbf{57.9} & 31.1 & 38.8 & 61.5 & 38.6 & 55.7 & 76.1 & 44.3 & 63.7 & 92.7 & 53.1 & 55.4 & 72.9 \\
\bottomrule
\end{tabular}%
}
\caption{Joint Hierarchical Extraction F1 performance for accident-type prediction and causal-factor extraction.}\label{tab:analysis-causal-factors}
% % \vspace{-1em}
\end{table*}

% These statistics show that \textsc{ConstructCIE} is challenging: models may perform well on frequent fields while struggling with rarer factors. We therefore evaluate models end-to-end, requiring accident-type prediction before causal extraction.

These distributions show substantial class imbalance across accident types, causal factors, and sub-causal factors. We therefore report category-level results in addition to aggregate end-to-end performance.

% \kh{We should mention there are two settings for testing here.}
% \kh{We are proposing a new task. What would be the evaluation metrics? We should introduce here.}
% \kh{Need to verify dist data}

\subsection{Evaluation Metrics}

 % We evaluate extraction performance using three span-matching strategies: exact string match, soft string match, and keyword-level match. We consider a factor prediction to be correct if each span in the prediction matches with an unique gold span.
 % \kh{Need to discuss why these three.}

% We use exact string match (E), soft string match (S), and keyword match (K) to evaluate correctness from strict span reproduction to core causal meaning. Each predicted span must match a unique gold span. A prediction is correct only if each predicted span is matched to a unique gold span.

We evaluate accident-type prediction and classification-based sub-causal factors using exact label matching and report F1. For extraction-based causal factors and sub-causal factors, we use exact string match (E), soft string match (S), and keyword match (K) to evaluate correctness from strict span reproduction to core causal meaning. Predicted and gold spans are matched one-to-one, so each span can participate in at most one match.

% Exact string match only considers a prediction to be correct if the predicted span exactly matches the annotated gold span. This strict metric measures complete span reproduction.

Exact string match requires the predicted span to exactly match the gold span.

Soft string match relaxes the exact-match requirement by measuring the similarity between the predicted and gold spans using the Gestalt string matching algorithm \cite{gestalt}, implemented with the \texttt{difflib} library. A prediction is considered correct if its similarity score is greater than or equal to the similarity threshold, 0.8. This allows minor boundary, wording, or formatting differences.

Keyword-level match evaluates whether the predicted span preserves the essential causal information in the gold annotation. Specifically, a prediction is considered correct if it contains all required keywords associated with the corresponding gold span. This metric allows limited variation in span boundaries, while ensuring that the core causal meaning of the annotation is retained.

\begin{table*}[t]
\centering
\scriptsize
\setlength{\tabcolsep}{2pt}
\renewcommand{\arraystretch}{0.95}
\resizebox{\textwidth}{!}{%
\begin{tabular}{l c c c c c c c c c c c c}
\toprule
\multirow{2}{*}{\textbf{Model}} 
& \multirow{2}{*}{\textbf{$k$}}
& \multicolumn{1}{c}{\textbf{Work. Circ.}}
& \multicolumn{2}{c}{\textbf{Mgmt.}}
& \multicolumn{2}{c}{\textbf{Cond.}}
& \multicolumn{2}{c}{\textbf{Equip.}}
& \multicolumn{2}{c}{\textbf{Behav.}}
& \multicolumn{2}{c}{\textbf{Conseq.}} \\
\cmidrule(lr){3-3}
\cmidrule(lr){4-5}
\cmidrule(lr){6-7}
\cmidrule(lr){8-9}
\cmidrule(lr){10-11}
\cmidrule(lr){12-13}
& 
& \shortstack[c]{\textbf{Constr.}\\\textbf{Trade}}
& \shortstack[c]{\textbf{Hazard}\\\textbf{Mgmt.}}
& \shortstack[c]{\textbf{Safety}\\\textbf{Train.}}
& \shortstack[c]{\textbf{Weather}\\\textbf{Cond.}}
& \shortstack[c]{\textbf{Workspace}\\\textbf{Cond.}}
& \shortstack[c]{\textbf{Prot.}\\\textbf{Equip.}}
& \shortstack[c]{\textbf{Work}\\\textbf{Equip.}}
& \shortstack[c]{\textbf{Inatt.}\\\textbf{Behav.}}
& \shortstack[c]{\textbf{Noncomp.}\\\textbf{Behav.}}
& \textbf{Severity}
& \shortstack[c]{\textbf{Body}\\\textbf{Part}} \\
\midrule
TagPrime-C & - & 69.7 & 47.8 & 40.0 & 44.2 & 18.0 & 32.4 & 44.1 & 38.0 & \textbf{61.6} & 95.4 & \textbf{86.0} \\
TagPrime-CR & - & 69.3 & 46.1 & 35.7 & 37.0 & 18.8 & 23.7 & \textbf{49.6} & 35.3 & 52.4 & 95.9 & 81.5 \\
\midrule
Llama3.2-3B & 0 & 0.0 & 0.0 & 0.0 & 0.0 & 0.0 & 0.0 & 0.0 & 0.0 & 0.0 & 36.7 & 27.8 \\
Llama3.2-3B & 5 & 27.2 & 0.0 & 18.1 & 0.0 & 6.3 & 0.0 & 2.3 & 5.3 & 5.7 & 69.9 & 66.6 \\
Llama3.2-3B & 30 & 33.5 & 0.0 & 12.4 & 2.5 & 11.2 & 0.0 & 3.6 & 14.1 & 14.4 & 81.3 & 48.5 \\
\midrule
Llama3.2-11B & 0 & 28.0 & 0.0 & 0.0 & 0.0 & 5.5 & 0.0 & 1.5 & 0.0 & 0.0 & 72.3 & 46.7 \\
Llama3.2-11B & 5 & 45.2 & 26.6 & 37.4 & 0.0 & 10.3 & 0.0 & 11.6 & 13.4 & 24.8 & 92.0 & 71.0 \\
Llama3.2-11B & 30 & 50.4 & 6.9 & 36.2 & 10.2 & 8.8 & 0.0 & 19.9 & 17.1 & 26.2 & 85.5 & 75.0 \\
\midrule
Llama3.1-70B & 0 & 54.8 & 20.2 & 0.0 & 4.0 & 4.4 & 0.0 & 16.8 & 29.3 & 5.7 & 90.1 & 75.3 \\
Llama3.1-70B & 5 & 68.9 & 24.0 & 25.7 & 20.6 & 18.5 & 15.2 & 31.0 & 31.6 & 43.1 & 95.4 & 77.9 \\
Llama3.1-70B & 30 & 77.4 & 30.6 & 27.9 & 45.1 & 22.9 & 12.4 & 35.0 & 31.8 & 46.4 & 92.9 & 81.8 \\
\midrule
Qwen3.5-9B & 0 & 53.3 & \textbf{48.0} & 25.0 & 17.0 & 6.4 & 0.0 & 5.2 & 36.0 & 35.9 & 81.5 & 65.8 \\
Qwen3.5-9B & 5 & 67.7 & 32.0 & 35.0 & 43.2 & 15.8 & 4.4 & 29.1 & 35.9 & 43.5 & 92.2 & 80.3 \\
Qwen3.5-9B & 30 & 70.8 & 19.9 & 50.0 & \textbf{54.3} & 19.4 & 5.7 & 39.5 & 29.9 & 42.2 & 95.2 & 82.1 \\
\midrule
Qwen3.5-27B & 0 & 60.7 & 35.9 & 50.0 & 18.7 & 7.3 & 17.2 & 37.5 & \textbf{45.5} & 24.5 & 90.0 & 60.4 \\
Qwen3.5-27B & 5 & 76.7 & 43.0 & 30.0 & 29.8 & 15.1 & \textbf{40.0} & 37.8 & 33.7 & 42.6 & 93.5 & 80.5 \\
Qwen3.5-27B & 30 & \textbf{79.0} & 42.0 & \textbf{55.0} & 35.7 & \textbf{23.2} & 13.0 & 47.1 & 37.7 & 47.5 & \textbf{97.9} & 81.4 \\
\bottomrule
\end{tabular}%
}
\caption{Joint Hierarchical Extraction exact-match F1 performance by sub-causal factor.}
\label{tab:e2ach-subfactor-f1}
% % \vspace{-1em}
\end{table*}

\section{Experiments}

\subsection{Models and Evaluation Strategy}

We evaluate large language models (LLMs) and supervised sequence-tagging models for the proposed CIE task. The LLMs include Llama-3.2-3B-Instruct (Llama3.2-3B),
Llama-3.2-11B-Vision-Instruct (Llama3.2-11B), Llama-3.1-70B-Instruct (Llama3.1-70B),
Qwen3.5-9B-Instruct (Qwen3.5-9B), and Qwen3.5-27B-Instruct (Qwen3.5-27B)
\cite{qwen35,llama31,llama32}. These models are evaluated using in-context learning, where the prompt asks the model to predict the accident type, extract causal factors and supporting spans, and then either classify or extract the corresponding sub-causal factor information. We evaluate in-context learning with
$k \in \{0,5,10,20,30,40\}$ examples to examine how performance changes
with additional demonstrations.

For supervised extraction, we use two TagPrime variants \cite{tagprime}: TagPrime with condition priming (TagPrime-C) and TagPrime with condition and relationship priming (TagPrime-CR). Because \textit{Object Involved} and \textit{Working Circumstances} may share the same text span, we extend TagPrime-C with an additional sequence-tagging head for \textit{Object Involved}.

All models are evaluated under the end-to-end setting, where the accident type is not provided and must first be predicted from the report. We compare two extraction strategies in this setting. End-to-End Joint Hierarchical Extraction (\textbf{JHE}) extracts all causal factors in a single grouped output and then extracts extraction-based sub-causal factors jointly while predicting classification-based sub-causal factors individually. End-to-End Individual Hierarchical Extraction (\textbf{IHE}) processes each causal factor separately by first determining whether it is present and then extracting or classifying its associated sub-causal factors.

For LLMs, the predicted accident type is also used to retrieve in-context examples: examples with the same predicted accident type are retrieved first, followed by examples from other accident types if needed. If no accident type can be predicted, examples are retrieved from all accident types. JHE and IHE therefore differ in how they organize causal-factor and sub-causal-factor extraction.

\subsection{Factor-Level Analysis}
% \kh{the subsection title should be Factor-Level}

% Tables~\ref{tab:analysis-causal-factors} and~\ref{tab:e2ach-subfactor-f1} report end-to-end Joint Hierarchical Extraction performance at the causal-factor and sub-causal-factor levels. Overall, the results show that current models can uncover broad causal meaning, but their performance is not yet sufficient for reliable construction-domain CIE. At the causal-factor level, Qwen models obtain strong keyword-level scores on several frequent and explicit fields. For example, Qwen3.5-9B reaches 94.0\% keyword F1 on \textit{Working Circumstances} and 74.6\% on \textit{Consequences} at $k=30$, while Qwen3.5-27B reaches 97.3\% keyword F1 on \textit{Behavioral Factors} and 86.4\% on \textit{Object Involved} at $k=30$.\

% Tables~\ref{tab:analysis-causal-factors} and~\ref{tab:e2ach-subfactor-f1} report end-to-end Joint Hierarchical Extraction performance at the causal-factor and sub-causal-factor levels. Overall, the evaluated models recover broad causal meaning but remain limited in precise extraction. At the causal-factor level, LLMs obtain high keyword F1 on several categories. For example, Qwen3.5-9B reaches 94.0\% keyword F1 on \textit{Working Circumstances} at $k=30$, while Qwen3.5-27B reaches 97.3\% keyword F1 on \textit{Behavioral Factors}. Llama3.1-70B is also competitive under exact matching, reaching 57.0\% on \textit{Working Circumstances} and 60.5\% on \textit{Consequences} at $k=30$.

Tables~\ref{tab:analysis-causal-factors} and~\ref{tab:e2ach-subfactor-f1} report end-to-end Joint Hierarchical Extraction performance at the accident-type, causal-factor, and sub-causal-factor levels. Overall, most evaluated models achieve strong accident-type prediction and recover broad causal meaning but remain limited in precise extraction. Qwen3.5-27B reaches 98.1\% accident-type F1 at $k=5$, while Llama3.2-3B improves from 8.7\% at $k=0$ to 64.9\% at $k=30$. At the causal-factor level, LLMs obtain high keyword F1 on several categories. For example, Qwen3.5-9B reaches 94.0\% keyword F1 on \textit{Working Circumstances} at $k=30$, while Qwen3.5-27B reaches 76.1\% on \textit{Behavioral Factors}. Llama3.1-70B is also competitive under exact matching, reaching 55.8\% on \textit{Working Circumstances} and 56.7\% on \textit{Object Involved} at $k=30$.

% The supervised TagPrime models remain competitive under exact matching. TagPrime-CR achieves 67.2\% exact F1 on \textit{Working Circumstances} and 85.9\% on \textit{Object Involved}, while TagPrime-C reaches 53.5\% on \textit{Equipment Factors} and 80.2\% on \textit{Behavioral Factors}. At the sub-causal-factor level, TagPrime-CR reaches 65.9\% on \textit{Protective Equipment Condition}. These results show that supervised sequence taggers remain strong for precise span extraction.

% The supervised TagPrime models remain competitive under exact matching. TagPrime-C obtains 86.4\% exact F1 on \textit{Object Involved} and 50.7\% on \textit{Equipment Factors}, while TagPrime-CR reaches 63.4\% on \textit{Working Circumstances} and 79.3\% on \textit{Behavioral Factors}. At the sub-causal-factor level, TagPrime-C reaches 61.6\% on \textit{Noncompliant Behavior} and 86.0\% on \textit{Affected Body Part}, while TagPrime-CR reaches 49.6\% on \textit{Work Equipment Condition}. These results show that supervised sequence taggers remain competitive for precise extraction.

The supervised TagPrime models remain competitive under exact matching. TagPrime-C obtains 86.4\% exact F1 on \textit{Managerial Factors} and 49.4\% on \textit{Equipment Factors}, while TagPrime-CR reaches 63.4\% on \textit{Working Circumstances} and 79.3\% on \textit{Consequences}. At the sub-causal-factor level, TagPrime-C reaches 61.6\% on \textit{Noncompliant Behavior} and 86.0\% on \textit{Affected Body Part}, while TagPrime-CR reaches 49.6\% on \textit{Work Equipment Condition}. These results show that supervised sequence taggers remain competitive for precise extraction.

% However, exact-match performance remains much lower than keyword-match performance, indicating that evaluated models often identify the correct causal content without reproducing the precise annotated span. For instance, Qwen3.5-9B at $k=30$ reaches 94.0\% keyword F1 on \textit{Working Circumstances}, but only 44.5\% exact F1. Thus, high keyword scores do not imply reliable evidence extraction.

% However, exact-match performance remains much lower than keyword-match performance for the LLMs, indicating that they often identify the correct causal content without reproducing the precise annotated span. For instance, Qwen3.5-9B at $k=30$ reaches 94.0\% keyword F1 on \textit{Working Circumstances}, but only 44.5\% exact F1. Similarly, Qwen3.5-27B reaches 97.3\% keyword F1 on \textit{Behavioral Factors}, but only 47.0\% exact F1. Thus, high keyword scores do not imply reliable evidence extraction.

However, exact-match performance remains much lower than keyword-match performance for the LLMs, indicating that they often identify the correct causal content without reproducing the precise annotated span. For instance, Qwen3.5-9B at $k=30$ reaches 94.0\% keyword F1 on \textit{Working Circumstances}, but only 46.0\% exact F1. Similarly, Qwen3.5-27B reaches 76.1\% keyword F1 on \textit{Behavioral Factors}, but only 38.6\% exact F1. Thus, high keyword scores do not imply reliable evidence extraction.

At the sub-causal-factor level, performance varies substantially across categories. At $k=30$, Qwen3.5-27B reaches 79.0\% exact F1 on \textit{Construction Trade}, 97.9\% on \textit{Severity}, and 81.4\% on \textit{Affected Body Part}. However, it reaches only 23.2\% on \textit{Workspace Condition} and 13.0\% on \textit{Protective Equipment Condition}. Llama3.1-70B similarly reaches 92.9\% on \textit{Severity} and 81.8\% on \textit{Affected Body Part}, but only 22.9\% on \textit{Workspace Condition} and 12.4\% on \textit{Protective Equipment Condition}. These results show substantial variation across sub-causal factors, with particularly low exact-match performance on \textit{Workspace Condition} and \textit{Protective Equipment Condition} for these models.

% For the stronger LLMs, \textit{Managerial Factors} and \textit{Working Condition Factors} remain among the most difficult causal-factor categories. These fields require grounding predictions to specific organizational deficiencies or site conditions, and their scores remain lower than those for more explicit fields. At the sub-causal-factor level, \textit{Severity}, \textit{Affected Body Part}, and \textit{Construction Trade} are relatively easy, while \textit{Workspace Condition}, \textit{Protective Equipment Condition}, and behavioral sub-causal factors remain difficult. For example, Qwen3.5-27B at $k=30$ reaches 97.9\% exact F1 on \textit{Severity}, but only 22.7\% on \textit{Workspace Condition} and 19.4\% on \textit{Protective Equipment Condition}.

% Few-shot prompting generally improves LLM performance, but gains vary across categories. Larger models achieve stronger exact extraction, while Qwen models remain especially competitive under keyword matching. Overall, evaluated supervised models remain competitive for precise span extraction, whereas larger LLMs recover broader causal meaning; however, low exact-match and rare-category performance still limit reliable use.

% Few-shot prompting generally improves LLM performance, but gains vary across models and categories. The supervised models remain competitive under exact matching, while the LLMs often obtain substantially higher keyword than exact F1. Thus, low exact-match performance and variation across categories continue to limit reliable extraction.

Few-shot prompting generally improves LLM performance, but gains vary across models and categories. The supervised models remain competitive under exact matching, while the LLMs often obtain substantially higher keyword than exact F1. Accident-type prediction is comparatively strong for most evaluated models, but low exact-match performance and variation across categories continue to limit reliable extraction.

\subsection{Few-Shot Scaling}

% Figure~\ref{fig:fewshot_scaling} shows that few-shot examples substantially improve LLM performance, with the largest gains occurring from $k=0$ to $k=5$. After this point, gains become smaller and are sometimes non-monotonic, indicating diminishing returns as more examples are added. Across exact, soft, and keyword matching, Qwen3.5-27B achieves the strongest overall performance, while Llama3.1-70B remains competitive under exact matching and Llama3.2-3B remains the weakest evaluated model.

\begin{figure*}[t]
    \centering
    \includegraphics[width=\textwidth]{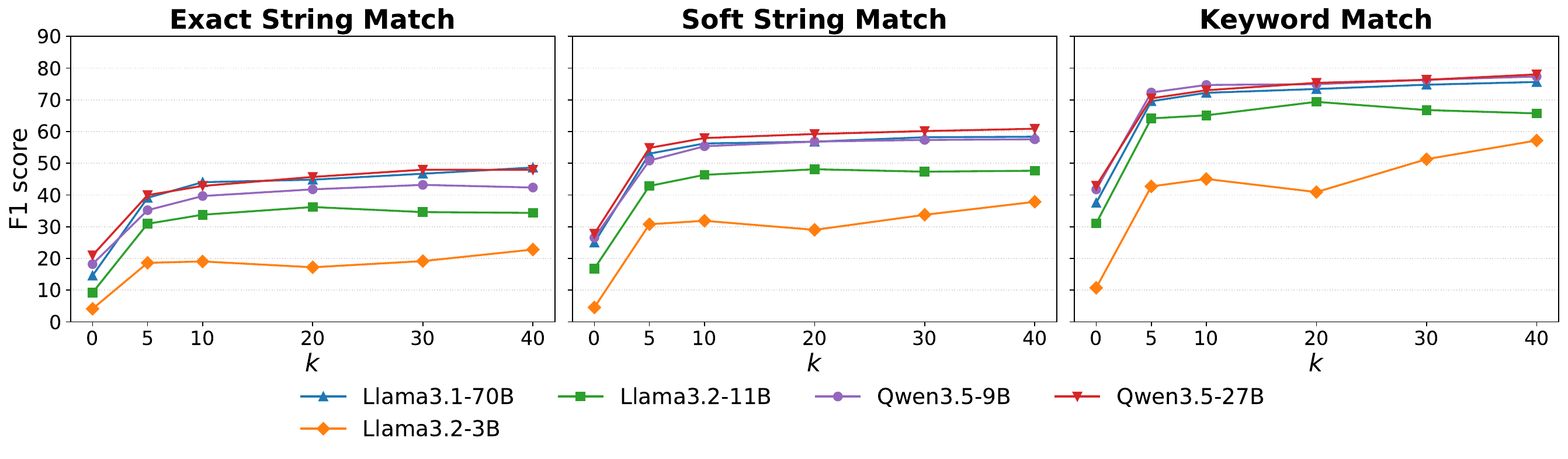}
    \caption{Average JHE F1 of LLMs under exact, soft, and keyword matching across in-context shot settings.}
    % \kh{update caption.}
    \label{fig:fewshot_scaling}
\end{figure*}

Figure~\ref{fig:fewshot_scaling} shows that few-shot examples substantially improve LLM performance, with the largest gains generally occurring from $k=0$ to $k=5$. After this point, gains become smaller and are sometimes non-monotonic, indicating diminishing returns. Across exact, soft, and keyword matching, Qwen3.5-27B achieves the strongest overall performance, while Llama3.1-70B remains competitive under exact matching and Llama3.2-3B remains the weakest evaluated model.

The scaling trend is consistent with the field-level results in Table~\ref{tab:analysis-causal-factors}: models generally benefit from additional demonstrations, although the magnitude of improvement varies by model and metric. Llama3.1-70B becomes competitive with Qwen3.5-27B under exact matching, whereas Qwen3.5-27B remains stronger overall under soft and keyword matching. Table~\ref{tab:e2ach-subfactor-f1} shows a similar pattern, where explicit fields such as \textit{Severity}, \textit{Affected Body Part}, and \textit{Construction Trade} are comparatively easier, while fine-grained categories such as \textit{Workspace Condition}, \textit{Protective Equipment Condition}, and managerial sub-causal factors remain difficult. Thus, few-shot prompting improves both semantic recovery and span extraction but does not fully resolve evidence-selection, boundary, and rare-category errors.

% The scaling trend is consistent with the field-level results in Table~\ref{tab:analysis-causal-factors}: models generally benefit from additional demonstrations, although the magnitude of improvement varies by model and metric. Llama3.1-70B becomes competitive with Qwen3.5-27B under exact matching, whereas Qwen3.5-27B remains stronger overall under soft and keyword matching. Table~\ref{tab:e2ach-subfactor-f1} also shows substantial variation across sub-causal factors. Thus, few-shot prompting improves overall performance but does not produce uniform gains across models, metrics, or categories.

\subsection{Strategy Comparison}

\begin{table}[t]
\centering
\small
\setlength{\tabcolsep}{4pt}
\begin{tabular}{l c ccc ccc}
\toprule
\multirow{2}{*}{\textbf{Model}}
& \multirow{2}{*}{\textbf{$k$}}
& \multicolumn{3}{c}{\textbf{JHE}}
& \multicolumn{3}{c}{\textbf{IHE}} \\
\cmidrule(lr){3-5} \cmidrule(lr){6-8}
&  & \textbf{E} & \textbf{S} & \textbf{K} & \textbf{E} & \textbf{S} & \textbf{K} \\
\midrule
Llama3.2-3B & 0 & 4.1 & 4.5 & 10.7 & 4.6 & 5.5 & 14.5 \\
Llama3.2-3B & 5 & 18.6 & 30.8 & 42.7 & 3.7 & 5.9 & 12.7 \\
Llama3.2-3B & 30 & 19.2 & 33.8 & 51.3 & 1.2 & 3.0 & 7.1 \\
\midrule
Llama3.2-11B & 0 & 9.2 & 16.8 & 31.1 & 7.0 & 15.9 & 27.1 \\
Llama3.2-11B & 5 & 30.9 & 42.9 & 64.2 & 17.8 & 28.6 & 52.9 \\
Llama3.2-11B & 30 & 34.6 & 47.4 & 66.8 & 23.4 & 33.4 & 63.0 \\
\midrule
Llama3.1-70B & 0 & 14.6 & 25.0 & 37.5 & 11.0 & 16.4 & 25.1 \\
Llama3.1-70B & 5 & 39.1 & 53.0 & 69.6 & 33.6 & 45.8 & 68.0 \\
Llama3.1-70B & 30 & 46.7 & 58.2 & 74.8 & 43.4 & 54.5 & 77.9 \\
\midrule
Qwen3.5-9B & 0 & 18.2 & 26.5 & 41.7 & 16.1 & 22.7 & 33.6 \\
Qwen3.5-9B & 5 & 35.2 & 50.9 & 72.3 & 34.1 & 46.3 & 74.2 \\
Qwen3.5-9B & 30 & 43.2 & 57.4 & 76.3 & 38.1 & 49.3 & 80.6 \\
\midrule
Qwen3.5-27B & 0 & 21.0 & 27.8 & 42.9 & 18.0 & 25.2 & 40.2 \\
Qwen3.5-27B & 5 & 40.0 & 54.9 & 70.5 & 36.9 & 48.1 & 73.3 \\
Qwen3.5-27B & 30 & 48.0 & 60.1 & 76.3 & 41.7 & 52.1 & 80.5 \\
\bottomrule
\end{tabular}
\caption{Comparison between joint and individual hierarchical extraction using average F1.}
\label{tab:analysis-strategy-comparison}
% \vspace{-1em}
\end{table}

% Table~\ref{tab:analysis-strategy-comparison} compares Joint Hierarchical Extraction (JHE) and Individual Hierarchical Extraction (IHE). Overall, JHE achieves stronger exact and soft matching across most models and shot settings, suggesting that extracting related causal factors in a grouped output provides more stable span-level evidence extraction. This advantage is especially clear for weaker models. For example, Llama3.2-3B, at $k=30$, reaches 26.9\% exact F1 and 55.9\% keyword F1 under JHE, compared with only 6.0\% exact F1 and 12.5\% keyword F1 under IHE. Llama3.2-11B shows a similar pattern, with JHE reaching 44.0\% exact F1 compared with 31.7\% under IHE.

% Table~\ref{tab:analysis-strategy-comparison} compares Joint Hierarchical Extraction (JHE) and Individual Hierarchical Extraction (IHE). Overall, JHE achieves stronger exact and soft matching across most models and shot settings, suggesting that extracting related causal factors in a grouped output provides more stable span-level evidence extraction. This advantage is especially clear for Llama3.2-3B. At $k=30$, JHE reaches 26.9\% exact F1 and 55.9\% keyword F1, compared with only 6.0\% exact F1 and 12.5\% keyword F1 under IHE. Llama3.2-11B shows a similar pattern, with JHE reaching 44.0\% exact F1 compared with 31.7\% under IHE.

Table~\ref{tab:analysis-strategy-comparison} compares Joint Hierarchical Extraction (JHE) and Individual Hierarchical Extraction (IHE). Overall, JHE achieves stronger exact and soft matching across most models and shot settings. This advantage is especially clear for Llama3.2-3B. At $k=30$, JHE reaches 19.2\% exact F1 and 51.3\% keyword F1, compared with only 1.2\% exact F1 and 7.1\% keyword F1 under IHE. Llama3.2-11B shows a similar pattern, with JHE reaching 34.6\% exact F1 compared with 23.4\% under IHE.

% For stronger models, the difference between JHE and IHE is smaller. JHE generally performs better under exact and soft matching, while IHE can achieve higher keyword F1. For example, Qwen3.5-27B at $k=30$ obtains 56.9\% exact and 67.8\% soft F1 with JHE, compared with 52.1\% exact and 60.1\% soft F1 with IHE. However, IHE reaches 82.0\% keyword F1, compared with 79.6\% for JHE. A similar pattern appears for Qwen3.5-9B, which reaches 82.0\% keyword F1 with IHE compared with 77.6\% with JHE.

% For the stronger evaluated models, the difference between JHE and IHE is smaller. JHE generally performs better under exact and soft matching, while IHE can achieve higher keyword F1. For example, Qwen3.5-27B at $k=30$ obtains 56.8\% exact and 67.6\% soft F1 with JHE, compared with 52.0\% exact and 60.0\% soft F1 with IHE. However, IHE reaches 82.0\% keyword F1, compared with 79.5\% for JHE. A similar pattern appears for Qwen3.5-9B, which reaches 81.9\% keyword F1 with IHE compared with 77.6\% with JHE. Llama3.1-70B also obtains stronger exact and soft performance with JHE but slightly higher keyword performance with IHE.

For the stronger evaluated models, the difference between JHE and IHE is smaller. JHE generally performs better under exact and soft matching, while IHE can achieve higher keyword F1. For example, Qwen3.5-27B at $k=30$ obtains 48.0\% exact and 60.1\% soft F1 with JHE, compared with 41.7\% exact and 52.1\% soft F1 with IHE. However, IHE reaches 80.5\% keyword F1, compared with 76.3\% for JHE. A similar pattern appears for Qwen3.5-9B, which reaches 80.6\% keyword F1 with IHE compared with 76.3\% with JHE. Llama3.1-70B also obtains stronger exact and soft performance with JHE but higher keyword performance with IHE.

% For stronger Qwen models, the difference between JHE and IHE is smaller. JHE generally performs better under exact and soft matching, while IHE sometimes achieves slightly higher keyword F1. For example, Qwen3.5-27B at $k=30$ obtains 56.9\% exact and 67.8\% soft F1 with JHE, compared with 52.1\% exact and 60.1\% soft F1 with IHE; however, IHE reaches a higher keyword F1 of 82.0\% compared with 79.6\%. Thus, IHE may improve semantic recovery, while JHE better supports precise extraction.

% This pattern is consistent with the few-shot trend in Figure~\ref{fig:fewshot_scaling} and the field-level results in Table~\ref{tab:analysis-causal-factors}. Although additional few-shot examples generally improve performance, the gains are model- and strategy-dependent. Exact-match performance remains substantially lower than keyword-match performance. We therefore use JHE as the main setting and use IHE to examine factor-wise semantic recovery.

% This pattern is consistent with the few-shot trend in Figure~\ref{fig:fewshot_scaling} and the field-level results in Table~\ref{tab:analysis-causal-factors}. Although additional few-shot examples generally improve performance, the gains are model- and strategy-dependent. Exact-match performance remains substantially lower than keyword-match performance. We therefore use JHE as the main extraction strategy and use IHE to examine factor-wise semantic recovery and strategy-dependent behavior.

This pattern is consistent with the few-shot trend in Figure~\ref{fig:fewshot_scaling} and the field-level results in Table~\ref{tab:analysis-causal-factors}. Although additional few-shot examples generally improve performance, the gains are model- and strategy-dependent. Exact-match performance remains substantially lower than keyword-match performance. We therefore use JHE as the main extraction strategy and use IHE to examine factor-wise semantic recovery and strategy-dependent behavior.

\subsection{Error Analysis}

% We further analyze extraction errors under exact matching across both Joint Hierarchical Extraction (JHE) and Individual Hierarchical Extraction (IHE) using six error types. \textbf{Over} refers to over-extraction, where the predicted span is a superset of the gold span. \textbf{Under} refers to under-extraction, where the prediction is a subset of the gold span. \textbf{Halluc.} denotes a prediction that cannot be matched to any substring of the source report using strict, case-insensitive, or normalized substring matching. \textbf{Joined} indicates that a prediction merges two or more separate gold spans into one span. \textbf{Mis-la} occurs when the prediction matches a gold span from another factor under the same parent category. Finally, \textbf{Mis-ex} refers to mis-extraction, where the normalized prediction appears in the source text but does not match any gold spans.

We further analyze extraction errors under exact matching using six error types. \textbf{Over} refers to over-extraction, where the predicted span contains the gold span but includes additional text. \textbf{Under} refers to under-extraction, where the prediction contains only part of the gold span. \textbf{Halluc.} denotes a prediction that cannot be matched to any substring of the source report using strict, case-insensitive, or normalized substring matching. \textbf{Joined} indicates that a prediction merges two or more separate gold spans. \textbf{Mis-la} occurs when the prediction matches a gold span assigned to another factor under the same parent category. Finally, \textbf{Mis-ex} refers to a source-grounded prediction that does not match any annotated gold span. For LLMs, errors are aggregated across all shot settings separately for JHE and IHE, while supervised models are evaluated under their single extraction setting.

Table~\ref{tab:error} shows that extraction strategy substantially affects the distribution of errors produced by the evaluated LLMs. Across the Llama models, hallucination accounts for a smaller share of JHE errors than IHE errors. The difference is especially large for Llama3.2-3B, for which hallucination accounts for 87.4\% of IHE errors and 28.8\% of JHE errors. The corresponding shares are 36.0\% and 19.4\% for Llama3.2-11B and 19.0\% and 14.9\% for Llama3.1-70B. Under JHE, the remaining Llama errors shift toward source-grounded failures. Mis-extraction accounts for 38.8\%, 35.3\%, and 33.3\% of JHE errors for Llama3.2-3B, Llama3.2-11B, and Llama3.1-70B, respectively. These results show that hallucination accounts for a smaller share of JHE errors for the evaluated Llama models, but evidence-selection and span-boundary errors remain common.

Qwen models remain strongly source-grounded under both strategies, with hallucination accounting for at most 3.7\% of their errors. Their errors are instead dominated by mis-extraction, which accounts for 43.4--49.6\% under JHE and 47.2--52.5\% under IHE. Mis-labeling also accounts for a smaller share of JHE errors for the Qwen models: the share changes from 18.7\% under IHE to 7.0\% under JHE for Qwen3.5-9B and from 17.6\% to 6.3\% for Qwen3.5-27B. However, over-extraction and under-extraction account for larger shares of JHE errors than IHE errors. The supervised TagPrime models similarly show no hallucination errors but remain affected by mis-extraction and boundary errors. Joined-span errors are rare across all models and strategies. Overall, the JHE error distributions shift away from unsupported or mislabeled evidence toward source-grounded but incorrect or imprecise evidence spans.

\begin{table}[t]
\centering
\scriptsize
\setlength{\tabcolsep}{2pt}
\renewcommand{\arraystretch}{1.05}
\resizebox{\columnwidth}{!}{%
\begin{tabular}{lrrrrrr}
\toprule
\textbf{Model}
& \textbf{Over}
& \textbf{Under}
& \textbf{Halluc.}
& \textbf{Joined}
& \textbf{Mis-la}
& \textbf{Mis-ex} \\
\midrule
\multicolumn{7}{l}{\textbf{Supervised Models}} \\
\midrule
TagPrime-C
& 17.1 & 27.8 & 0.0 & 0.9 & 10.3 & 44.0 \\
TagPrime-CR
& 20.5 & 36.0 & 0.0 & 1.8 & 9.7 & 32.0 \\
\midrule
\multicolumn{7}{l}{\textbf{Joint Hierarchical Extraction (JHE)}} \\
\midrule
Llama3.2-3B
& 11.7 & 14.6 & 28.8 & 0.2 & 5.8 & 38.8 \\
Llama3.2-11B
& 20.4 & 16.6 & 19.4 & 0.6 & 7.6 & 35.3 \\
Llama3.1-70B
& 23.1 & 20.6 & 14.9 & 1.0 & 7.1 & 33.3 \\
Qwen3.5-9B
& 25.6 & 19.1 & 3.7 & 1.2 & 7.0 & 43.4 \\
Qwen3.5-27B
& 21.3 & 19.4 & 2.4 & 1.0 & 6.3 & 49.6 \\
\midrule
\multicolumn{7}{l}{\textbf{Individual Hierarchical Extraction (IHE)}} \\
\midrule
Llama3.2-3B
& 1.2 & 2.8 & 87.4 & 0.1 & 1.0 & 7.5 \\
Llama3.2-11B
& 10.5 & 8.3 & 36.0 & 0.3 & 8.8 & 36.0 \\
Llama3.1-70B
& 11.7 & 16.4 & 19.0 & 0.6 & 14.9 & 37.5 \\
Qwen3.5-9B
& 15.8 & 14.4 & 3.2 & 0.7 & 18.7 & 47.2 \\
Qwen3.5-27B
& 13.5 & 13.8 & 2.0 & 0.6 & 17.6 & 52.5 \\
\bottomrule
\end{tabular}}
\caption{Exact-match error distributions by extraction strategy.}
\label{tab:error}
\end{table}

\section{Discussion}

The experimental results show that construction-domain CIE remains challenging for the evaluated models. Most evaluated models perform strongly on accident-type prediction, indicating that they can identify the overall accident mechanism from the narrative. Across causal-factor and sub-causal-factor results, the evaluated models perform better on frequent and explicit information, such as \textit{Working Circumstances}, \textit{Consequences}, \textit{Construction Trade}, \textit{Severity}, and \textit{Affected Body Part}, but struggle with more fine-grained causal evidence, such as \textit{Workspace Condition}, \textit{Protective Equipment Condition}, and several managerial and behavioral sub-causal factors. For the evaluated models, identifying high-level accident information therefore appears easier than selecting the precise textual evidence explaining why an accident occurred.

A consistent pattern across the evaluated models is the gap between keyword and exact matching. Higher keyword F1 indicates that the models often identify the central causal content of the annotation, even when the predicted span does not fully match the gold evidence. In contrast, lower exact F1 shows that the evaluated models frequently fail to reproduce the complete annotated span with the required boundaries. The error analysis helps explain this difference. Mis-extraction may preserve relevant causal information while selecting the wrong source-supported span, whereas over-extraction includes unnecessary surrounding context and under-extraction omits part of the required evidence. These errors may still retain keywords associated with the annotated causal content while failing strict span-level evaluation. Thus, for the evaluated models, stronger source grounding and broad causal recovery do not necessarily lead to precise evidence extraction. Accurate accident-type prediction likewise does not eliminate downstream evidence-selection and span-boundary errors.

The comparison between JHE and IHE suggests a trade-off between joint and individual extraction for the evaluated models. JHE generally achieves stronger exact and soft matching, while IHE sometimes achieves higher keyword F1. The error distributions further show that unsupported or mislabeled predictions account for smaller shares of JHE errors in several cases, while over-extraction and under-extraction account for larger shares. Because precise and source-grounded evidence extraction is central to CIE, JHE is more suitable as the main extraction strategy in this work.

% Overall, these findings indicate that construction accident CIE requires more than general language understanding. Models must handle domain-specific terminology, long-span evidence, class imbalance, and implicit causal relations. Future work should improve span grounding, incorporate construction-domain knowledge, and better handle rare causal categories.

Overall, these findings indicate that the evaluated models require more than general language understanding for construction accident CIE. They must handle domain-specific terminology, long-span evidence, class imbalance, implicit causal relations, and precise evidence boundaries. Future work should improve span grounding, incorporate construction-domain knowledge, and better handle rare causal categories.

\section{Conclusion}

We introduced \textsc{ConstructCIE}, a construction-specific dataset with annotations for accident types, causal factors, sub-causal factors, and supporting evidence from OSHA accident narratives. End-to-end experiments show that most evaluated models predict accident type accurately and recover broad causal meaning but remain limited in precise evidence extraction. JHE generally achieves stronger exact and soft matching, while IHE can achieve higher keyword performance. Error distributions vary by extraction strategy, but evidence-selection and span-boundary errors remain common. Reliable construction accident CIE therefore requires accurate high-level accident understanding together with more precise and better-grounded evidence extraction.

\section*{Authors' Contributions}

Jaehoon Lee helped curate the dataset and served as one of two construction-safety annotators. Hung Nguyen designed the evaluation, preprocessed the data, identified quality issues, conducted the experiments and error analysis, and wrote the manuscript. Namgyun Kim and Kuan-Hao Huang advised the project.
% Portions of this research were conducted with the advanced computing resources provided by Texas A\&M High Performance Research Computing.

\section*{Acknowledgments}

We thank the anonymous reviewers for their valuable feedback and constructive comments. Portions of this research were conducted with the advanced computing resources provided by Texas A\&M High Performance Research Computing.

\section*{Limitations}

This work is limited by computational resources. The largest LLM evaluated was Llama3.1-70B, and we did not test larger open-weight or closed-source models. Therefore, the results characterize the evaluated models rather than an upper bound on CIE performance. In addition, \textsc{ConstructCIE} is manually annotated from OSHA accident narratives, which provides domain-informed labels but limits dataset scale. Some important causal categories, such as managerial factors, safety training deficiencies, weather conditions, and protective equipment conditions, are relatively rare, creating class imbalance. Finally, our experiments focus on text-only end-to-end extraction and do not incorporate external construction safety knowledge or multimodal project evidence. Future work should expand the dataset, evaluate larger models, and improve domain-grounded span extraction.

% Bibliography entries for the entire Anthology, followed by custom entries
%\bibliography{anthology,custom}
% Custom bibliography entries only
% \newpage
\bibliography{custom}

\clearpage

\appendix

\section{Causal and Sub-causal Factors Definitions}
\label{sec:extraction_framework}
The semantic framework categorizes causality into seven causal factors and their associated sub-causal factors:

\begin{itemize}
    \item \textbf{Working Circumstances}: Physical and operational contexts present at the time of the accident that characterize the immediate work situation.
    \begin{itemize}
        \item \textbf{Construction Trade}: The primary construction activity division being performed at the time of the accident.
        
    \end{itemize}
    
    \item \textbf{Object Involved}: The equipment, structure, material, or object that physically interacted with or contributed to the accident mechanism.

    \item \textbf{Managerial Factors}: Deficiencies at the organizational or supervisory level that allowed unsafe conditions or behaviors to exist.
    \begin{itemize}
        \item \textbf{Failure of Hazard Management}: Failure to identify or include foreseeable hazards during pre-task planning or risk assessment.
        \item \textbf{Deficiency in Safety Training}: Failure to provide adequate task-specific training enabling workers to perform tasks safely and respond to foreseeable hazards.
    \end{itemize}

    \item \textbf{Working Condition Factors}: Physical conditions of the work environment, such as natural conditions and workspace characteristics, that contributed to the occurrence of the accident.
    \begin{itemize}
        \item \textbf{Weather Condition}: Weather-related conditions that influenced the work environment or contributed to the accident sequence.
        \item \textbf{Workspace Condition}: Physical conditions related to the workspace, or structural elements, including spatial arrangement, support conditions, integrity, surface characteristics, or surrounding physical layout.
    \end{itemize}

    \item \textbf{Equipment Factors}: Conditions related to personal or collective protective equipment or work equipment whose state contributed to the accident sequence.
    \begin{itemize}
        \item \textbf{Protective Equipment Condition}: The state, functionality, availability, configuration, or appropriateness of personal or collective protective equipment.
        \item \textbf{Work Equipment Condition}: The functional state, physical integrity, configuration, or availability of tools, machinery, or powered/non-powered work equipment.
    \end{itemize}

    \item \textbf{Behavioral Factors}: Task-level human factors involving cognitive lapses or procedural deviations that directly contributed to the occurrence of the accident.
    \begin{itemize}
        \item \textbf{Inattentive Behavior}: Behavior resulting from cognitive lapses such as reduced vigilance, distraction, or loss of situational awareness.
        \item \textbf{Noncompliant Behavior}: Behavior involving the disregard of required safety rules, safe work practices, or protective measures.
    \end{itemize}

    \item \textbf{Consequences}: The adverse outcomes of the incident, including the seriousness of injury and the part(s) of the body affected.
    \begin{itemize}
        \item \textbf{Severity}: The degree and seriousness of injury, ranging from minor to fatal.
        \item \textbf{Affected Body Part}: The part(s) of the body that sustained harm due to the incident.
    \end{itemize}
\end{itemize}

\section{Supervised Model Implementation Details}

\subsection{TagPrime}

% TagPrime
TAGPRIME \cite{tagprime} is a sequence-tagging model that uses condition and relation priming to make token representations task-specific for relational structure extraction. For classification tasks, we fine-tune the encoder using the contextualized \texttt{[CLS]} representation, which is passed through a linear classification head and optimized with cross-entropy loss. We run our experiments on an NVIDIA A100 machine. The major hyperparameters for this model's variant TagPrime-C and TagPrime-CR are listed in Table~\ref{tab:tagc} and Table~\ref{tab:tagcr}, respectively. 

\begin{table}[htbp]
    \centering
    \begin{tabular}{ll}
        \toprule
        \textbf{Parameter} & \textbf{Value} \\
        \midrule
        Pretrained Model Name & RoBERTa-large \\
        Max Length & 512 \\
        Train Batch Size & 4 \\
        Eval Batch Size & 8 \\
        Accumulate Step & 1 \\
        Grad Clipping & 1.0 \\
        Base Model Learning Rate & 2e-5 \\
        Base Model Weight Decay & 1e-5 \\
        Learning Rate & 1e-4 \\
        Weight Decay & 1e-5 \\
        Warmup Epoch & 3 \\
        Max Epoch & 50 \\
        Base Model Dropout & 0.1 \\
        Linear Dropout & 0.4 \\
        Linear Hidden Num & 150 \\
        Linear Bias & True \\
        Linear Activation & relu \\
        Multi Piece Strategy & average \\
        Use CRF & True \\
        Priming Type & condition \\
        \bottomrule
    \end{tabular}
    \caption{TagPrime-C Hyperparameter details.}
    \label{tab:tagc}
\end{table}

\begin{table}[htbp]
    \centering
    \begin{tabular}{ll}
        \toprule
        \textbf{Parameter} & \textbf{Value} \\
        \midrule
        Pretrained Model Name & RoBERTa-large \\
        Max Length & 512 \\
        Train Batch Size & 4 \\
        Eval Batch Size & 8 \\
        Accumulate Step & 1 \\
        Grad Clipping & 1.0 \\
        Base Model Learning Rate & 2e-5 \\
        Base Model Weight Decay & 1e-5 \\
        Learning Rate & 1e-4 \\
        Weight Decay & 1e-5 \\
        Warmup Epoch & 3 \\
        Max Epoch & 50 \\
        Base Model Dropout & 0.1 \\
        Linear Dropout & 0.4 \\
        Linear Hidden Num & 150 \\
        Linear Bias & True \\
        Linear Activation & relu \\
        Multi Piece Strategy & average \\
        Use CRF & True \\
        Priming Type & condition+relation \\
        \bottomrule
    \end{tabular}
    \caption{TagPrime-CR Hyperparameter details.}
    \label{tab:tagcr}
\end{table}

\section{Causal and Sub-causal Factor Examples}

Figures~\ref{fig:appendix-example-annotation-1}--\ref{fig:appendix-example-annotation-5} show representative \textsc{ConstructCIE} annotations covering all accident types and causal-factor categories. Each example presents the accident narrative followed by its annotated hierarchy, where causal factors are shown in bold and sub-causal factors are marked with $\hookrightarrow$.

% Selected 5 records to cover all targets.
%   id=154295.015 type=fall covers=['working_circumstances', 'object_involved', 'working_condition_factors', 'equipment_factors', 'behavioral_factors', 'consequences', 'construction_trade', 'workspace_condition', 'work_equipment_condition', 'noncompliant_behavior', 'severity', 'affected_body_part']
%   id=128765.015 type=caught-in/between covers=['working_circumstances', 'object_involved', 'managerial_factors', 'consequences', 'construction_trade', 'failure_of_hazard_management', 'deficiency_in_safety_training', 'severity']
%   id=125370.015 type=struck-by covers=['working_circumstances', 'object_involved', 'behavioral_factors', 'construction_trade', 'inattentive_behavior']
%   id=97346.015 type=electrocution covers=['working_circumstances', 'working_condition_factors', 'consequences', 'construction_trade', 'weather_condition', 'severity']
%   id=106271.015 type=fall covers=['working_circumstances', 'equipment_factors', 'consequences', 'construction_trade', 'protective_equipment_condition', 'severity']

\begin{figure*}[t]
\centering
\small
\setlength{\fboxsep}{8pt}
\fbox{
\begin{minipage}{0.95\textwidth}

\textbf{Accident Report:}
An employee working for a painting contractor had positioned an extension ladder up against a roof facia/eve approximately 16 feet above the ground. The employee then climbed the ladder with a paint roller in his hand and was standing on the ladder approximately eleven feet above the ground. The employee was using the paint roller to paint the facia/roof eve on a two-story single-family residence which had been damaged by a hurricane. The employee was using a ladder owned by the homeowner which he found at the rear of the property. The ladder was missing the feet and the employee had set it up with the ladder base six feet from the base of the building in violation of the 4:1 rule. The employee also rested the ladder on a painted concrete surface which reduced the coefficient of friction of the concrete. The ladder slipped out from under the employee causing him to fall to the concrete ground below. The employee died from severe head injuries.

\vspace{0.8em}

\renewcommand{\arraystretch}{1.25}
\begin{tabularx}{\textwidth}{|p{0.32\textwidth}|X|}
\hline
\multicolumn{2}{|c|}{\textbf{Accident Type: Fall}} \\
\hline

\textbf{Working Circumstances}
& An employee working for a painting contractor had positioned an extension ladder up against a roof facia/eve approximately 16 feet above the ground. The employee then climbed the ladder with a paint roller in his hand and was standing on the ladder approximately eleven feet above the ground. The employee was using the paint roller to paint the facia/roof eve on a two-story single-family residence which had been damaged by a hurricane. \\
\hline

\quad $\hookrightarrow$ \textbf{Construction Trade}
& Finishes \\
\hline

\textbf{Object Involved}
& ladder \\
\hline

\textbf{Working Condition Factors}
& The employee also rested the ladder on a painted concrete surface which reduced the coefficient of friction of the concrete \\
\hline

\quad $\hookrightarrow$ \textbf{Workspace Condition}
& rested the ladder on a painted concrete surface which reduced the coefficient of friction of the concrete \\
\hline

\textbf{Equipment Factors}
& The ladder was missing the feet \\
\hline

\quad $\hookrightarrow$ \textbf{Work Equipment Condition}
& ladder was missing the feet \\
\hline

\textbf{Behavioral Factors}
& the employee had set it up with the ladder base six feet from the base of the building in violation of the 4:1 rule \\
\hline

\quad $\hookrightarrow$ \textbf{Noncompliant Behavior}
& had set it up with the ladder base six feet from the base of the building in violation of the 4:1 rule \\
\hline

\textbf{Consequences}
& The employee died from severe head injuries \\
\hline

\quad $\hookrightarrow$ \textbf{Severity}
& Fatality \\
\hline

\quad $\hookrightarrow$ \textbf{Affected Body Part}
& head \\
\hline

\end{tabularx}

\end{minipage}
}
\caption{Fall example illustrating \textit{Working Circumstances}, \textit{Object Involved}, \textit{Working Condition Factors}, \textit{Equipment Factors}, \textit{Behavioral Factors}, and \textit{Consequences}.}
\label{fig:appendix-example-annotation-1}
\end{figure*}

\begin{figure*}[t]
\centering
\small
\setlength{\fboxsep}{8pt}
\fbox{
\begin{minipage}{0.95\textwidth}

\textbf{Accident Report:}
An employee was working as a construction laborer for a landscaping service. He and some coworkers were working in an excavation. They were installing 150-millimeter (6-inch) drain line for an underground utilities project. The excavation had been dug by their foreman, who operated the equipment. The excavation was approximately 8.2 meters (27 feet) in length by 3.7 meters (12 feet) in depth. The angle of repose was approximately 88 to 90 degrees on the north, south, and east sides. Soil conditions consisted of sand mixed with clay and some gravel, with previously excavated material, consisting of varying levels of sand and clay, mixed in at different depths. Fissures developed in the excavation walls as the depth of the excavation increased. The workers did not immediately report the hazardous condition of the excavation walls, as none of them recognized the hazard associated with changing soil conditions. No accident prevention program had been developed or coordinated with the workers. The foreman and equipment operator was working at the jobsite and was in direct and plain view of the workers exposed to the excavation collapse hazard. The near-vertical walls collapsed onto the employee. He suffered multiple severe injuries, and he was asphyxiated by being buried under excavated materials. He was killed. At one place, the account says that he was hospitalized, so he might have died after being removed from the collapse.

\vspace{0.8em}

\renewcommand{\arraystretch}{1.25}
\begin{tabularx}{\textwidth}{|p{0.32\textwidth}|X|}
\hline
\multicolumn{2}{|c|}{\textbf{Accident Type: Caught-in/between}} \\
\hline

\textbf{Working Circumstances}
& An employee was working as a construction laborer for a landscaping service. He and some coworkers were working in an excavation. They were installing 150-millimeter (6-inch) drain line for an underground utilities project. \\
\hline

\quad $\hookrightarrow$ \textbf{Construction Trade}
& Utilities \\
\hline

\textbf{Object Involved}
& walls \\
\hline

\textbf{Managerial Factors}
& none of them recognized the hazard associated with changing soil conditions. No accident prevention program had been developed or coordinated with the workers \\
\hline

\quad $\hookrightarrow$ \textbf{Failure of Hazard Management}
& none of them recognized the hazard associated with changing soil conditions \\
\hline

\quad $\hookrightarrow$ \textbf{Deficiency in Safety Training}
& No accident prevention program had been developed or coordinated with the workers \\
\hline

\textbf{Consequences}
& He suffered multiple severe injuries, and he was asphyxiated by being buried under excavated materials. He was killed \\
\hline

\quad $\hookrightarrow$ \textbf{Severity}
& Fatality \\
\hline

\end{tabularx}

\end{minipage}
}
\caption{Caught-in/between example illustrating \textit{Working Circumstances}, \textit{Object Involved}, \textit{Managerial Factors}, and \textit{Consequences}.}
\label{fig:appendix-example-annotation-2}
\end{figure*}

\begin{figure*}[t]
\centering
\small
\setlength{\fboxsep}{8pt}
\fbox{
\begin{minipage}{0.95\textwidth}

\textbf{Accident Report:}
An employee working as a carpenter for a residential remodeler was working on a residential construction site. The employer has only 1 to 2 employees. The employee was repairing side panels on the south side of a house on a narrow walkway. A scaffold had been erected in the walkway, which was approximately three feet wide. The employee was using a nail gun when he stood up with his finger on the trigger. The employee inadvertently hit the trigger causing the nail gun to eject a nail into his left leg. The employee was experienced using nail guns. He was treated at the hospital and released on the same day.

\vspace{0.8em}

\renewcommand{\arraystretch}{1.25}
\begin{tabularx}{\textwidth}{|p{0.32\textwidth}|X|}
\hline
\multicolumn{2}{|c|}{\textbf{Accident Type: Struck-by}} \\
\hline

\textbf{Working Circumstances}
& An employee working as a carpenter for a residential remodeler was working on a residential construction site. The employer has only 1 to 2 employees. The employee was repairing side panels on the south side of a house on a narrow walkway. A scaffold had been erected in the walkway, which was approximately three feet wide. The employee was using a nail gun when he stood up with his finger on the trigger. \\
\hline

\quad $\hookrightarrow$ \textbf{Construction Trade}
& Wood, plastics, and composites \\
\hline

\textbf{Object Involved}
& nail \\
\hline

\textbf{Behavioral Factors}
& The employee inadvertently hit the trigger causing the nail gun to eject a nail \\
\hline

\quad $\hookrightarrow$ \textbf{Inattentive Behavior}
& inadvertently hit the trigger \\
\hline

\end{tabularx}

\end{minipage}
}
\caption{Struck-by example illustrating \textit{Working Circumstances}, \textit{Object Involved}, and \textit{Behavioral Factors}.}
\label{fig:appendix-example-annotation-3}
\end{figure*}

\begin{figure*}[t]
\centering
\small
\setlength{\fboxsep}{8pt}
\fbox{
\begin{minipage}{0.95\textwidth}

\textbf{Accident Report:}
Employees \#1, \#2, and \#3, employed by a roofing company, were engaged in roofing work at a residential building construction site. They were installing roof shingles when lightning struck the roof and made contact with all three employees. Emergency services were called to transport the injured employees to the hospital for treatment of electrical shock and burns. The subsequent investigation report indicated that Employees \#1 and \#2 required admission and hospitalization.

\vspace{0.8em}

\renewcommand{\arraystretch}{1.25}
\begin{tabularx}{\textwidth}{|p{0.32\textwidth}|X|}
\hline
\multicolumn{2}{|c|}{\textbf{Accident Type: Electrocution}} \\
\hline

\textbf{Working Circumstances}
& Employees \#1, \#2, and \#3, employed by a roofing company, were engaged in roofing work at a residential building construction site. They were installing roof shingles \\
\hline

\quad $\hookrightarrow$ \textbf{Construction Trade}
& Thermal and moisture protection \\
\hline

\textbf{Working Condition Factors}
& lightning struck the roof \\
\hline

\quad $\hookrightarrow$ \textbf{Weather Condition}
& lightning struck the roof \\
\hline

\textbf{Consequences}
& electrical shock and burns \\
\hline

\quad $\hookrightarrow$ \textbf{Severity}
& Hospitalized injury \\
\hline

\end{tabularx}

\end{minipage}
}
\caption{Electrocution example illustrating \textit{Working Circumstances}, \textit{Working Condition Factors}, and \textit{Consequences}.}
\label{fig:appendix-example-annotation-4}
\end{figure*}

\begin{figure*}[t]
\centering
\small
\setlength{\fboxsep}{8pt}
\fbox{
\begin{minipage}{0.95\textwidth}

\textbf{Accident Report:}
Employee \#1 was engaged in roofing work at a two-story residence. He was applying shingles when he fell and struck the hard dirt surface, a fall height of 20 feet. Emergency services were called, and Employee \#1 was transported to the hospital. He was admitted and treated for fall-related injuries. The investigation determined that at some point during the work, the fall protection system Employee \#1 was using failed.

\vspace{0.8em}

\renewcommand{\arraystretch}{1.25}
\begin{tabularx}{\textwidth}{|p{0.32\textwidth}|X|}
\hline
\multicolumn{2}{|c|}{\textbf{Accident Type: Fall}} \\
\hline

\textbf{Working Circumstances}
& Employee \#1 was engaged in roofing work at a two-story residence. He was applying shingles \\
\hline

\quad $\hookrightarrow$ \textbf{Construction Trade}
& Thermal and moisture protection \\
\hline

\textbf{Equipment Factors}
& the fall protection system Employee \#1 was using failed \\
\hline

\quad $\hookrightarrow$ \textbf{Protective Equipment Condition}
& fall protection system Employee \#1 was using failed \\
\hline

\textbf{Consequences}
& He was admitted and treated for fall-related injuries \\
\hline

\quad $\hookrightarrow$ \textbf{Severity}
& Hospitalized injury \\
\hline

\end{tabularx}

\end{minipage}
}
\caption{Fall example illustrating \textit{Working Circumstances}, \textit{Equipment Factors}, and \textit{Consequences}.}
\label{fig:appendix-example-annotation-5}
\end{figure*}

\section{Prompts for LLMs}
\label{app:prompts}

% Table~\ref{tab:prompt-jhe} illustrates the prompt we use for Joint Hierarchical Extraction (JHE), which lists all causal factors in a single grouped prompt so the model outputs one line per factor. Table~\ref{tab:prompt-ihe} illustrates the prompt for Individual Hierarchical Extraction (IHE), which issues one prompt per causal factor (shown here for \textit{Equipment Factors}). Table~\ref{tab:prompt-cls} illustrates the prompt for classification.

Table~\ref{tab:prompt-jhe} illustrates sub-causal-factor extraction under JHE, where related sub-causal factors are predicted jointly, using \textit{Weather Condition} and \textit{Workspace Condition} as an example. Table~\ref{tab:prompt-ihe} illustrates IHE, where a sub-causal factor is predicted separately, using \textit{Weather Condition} as an example. Table~\ref{tab:prompt-cls} illustrates the classification prompt using \textit{Severity} and the same format is used for accident-type classification.

\begin{table*}[t!]
\centering
\setlength{\tabcolsep}{5pt}
\resizebox{.99\textwidth}{!}{
\begin{tabular}{cl}
    \toprule
    \multicolumn{2}{l}{\textbf{Prompt Used for Extraction in JHE}}\\
    \midrule
    Instruction & \makecell[l]{
    You are a strict safety accident report analyzer. \\
    Your response must ONLY be a list of key-value pairs formatted as ``Alias: Value''. \\
    If information is missing, output ``None''. \\
    Task: Extract the exact, concise text spans from the Input Text that describe the following factors. \\
    - Weather Condition: Weather-related conditions that influenced the work environment or contributed \\
    to the accident sequence. \\
    - Workspace Condition: Physical conditions related to the workspace, or structural elements, \\
    including spatial arrangement, support conditions, integrity, surface characteristics, or \\
    surrounding physical layout. \\
    Your output MUST use the exact Alias provided below as the Key. \\
    Format: ``Alias: Extracted\_Span'' \\
    Example: If the factor is ``- object: ...'', your output MUST be ``object: bucket truck''. \\
    If there are multiple distinct occurrences for a single factor, separate them with a semicolon (;). \\
    Example: ``object: bucket truck; safety harness'' \\
    If a factor is not explicitly described, output ``None''. \\
    Output ONLY a single line per factor in the format ``Alias: Extracted\_Span''.}\\
    \\[-0.8em]
    Example 1 & \makecell[l]{
    Example 1: \\
    Text: gust of wind came through; set trusses were not braced in accordance with the building \\
    component safety information \\
    Weather Condition: gust of wind came through \\
    Workspace Condition: trusses were not braced in accordance with the building component safety \\
    information}\\
    \\[-0.8em]
    ... & ...\\
    \\[-0.8em]
    Query & \makecell[l]{
    Question: \\
    Text: unprotected leading edge of an elevated floor deck; wind gust caught the plywood}\\
    \\[-0.8em]
    Output & \makecell[l]{
    Weather Condition: wind gust caught the plywood \\
    Workspace Condition: unprotected leading edge of an elevated floor deck}\\
    \\[-0.8em]
    \bottomrule
\end{tabular}}
\caption{Prompt used for Joint Hierarchical Extraction.}
\label{tab:prompt-jhe}
\end{table*}

\begin{table*}[t!]
\centering
\setlength{\tabcolsep}{5pt}
\resizebox{.99\textwidth}{!}{
\begin{tabular}{cl}
    \toprule
    \multicolumn{2}{l}{\textbf{Prompt Used for Extraction in IHE}}\\
    \midrule
    Instruction & \makecell[l]{
    You are a strict safety accident report analyzer. \\
    Your response must ONLY be a list of key-value pairs formatted as ``Alias: Value''. \\
    If information is missing, output ``None''. \\
    Task: Extract the exact, concise text spans from the Input Text that describe the following factors. \\
    - Weather Condition: Weather-related conditions that influenced the work environment or contributed \\
    to the accident sequence. \\
    Your output MUST use the exact Alias provided below as the Key. \\
    Format: ``Alias: Extracted\_Span'' \\
    Example: If the factor is ``- object: ...'', your output MUST be ``object: bucket truck''. \\
    If there are multiple distinct occurrences for a single factor, separate them with a semicolon (;). \\
    Example: ``object: bucket truck; safety harness'' \\
    If a factor is not explicitly described, output ``None''. \\
    Output ONLY a single line per factor in the format ``Alias: Extracted\_Span''.}\\
    \\[-0.8em]
    Example 1 & \makecell[l]{
    Example 1: \\
    Text: gust of wind came through; set trusses were not braced in accordance with the building \\
    component safety information \\
    Weather Condition: gust of wind came through}\\
    \\[-0.8em]
    ... & ...\\
    \\[-0.8em]
    Query & \makecell[l]{
    Question: \\
    Text: unprotected leading edge of an elevated floor deck; wind gust caught the plywood}\\
    \\[-0.8em]
    Output & \makecell[l]{
    Weather Condition: wind gust caught the plywood}\\
    \\[-0.8em]
    \bottomrule
\end{tabular}}
\caption{Prompt used for Individual Hierarchical Extraction.}
\label{tab:prompt-ihe}
\end{table*}

\begin{table*}[t!]
\centering
\setlength{\tabcolsep}{5pt}
\resizebox{.99\textwidth}{!}{
\begin{tabular}{cl}
    \toprule
    \multicolumn{2}{l}{\textbf{Prompt Used for Classification}}\\
    \midrule
    Instruction & \makecell[l]{
    You are a strict safety accident report analyzer. Your response must ONLY be a list of key-value pairs \\
    formatted as ``Alias: Value''. If information is missing, output ``None''.\\
    Task: Classify the following factors based on the input text into exactly one of the provided \\
    allowed classes:\\
    - Severity: The degree and seriousness of injury, ranging from minor to fatal. (Allowed Classes: \\
    fatality, hospitalized injury, non hospitalized injury)\\
    Your output MUST use the exact Alias provided above as the Key.\\
    Format: ``Alias: Selected\_Class''\\
    Example: If the factor is ``- task: ...'', your output must start with ``task: '' followed by the class.}\\
    \\[-0.8em]
    Example 1 & \makecell[l]{
    Example 1:\\
    Text: Employee \#1 was removing concrete forms at the edge of an excavation when the wall collapsed, \\
    burying him. He was pronounced dead at the scene.\\
    Severity: fatality}\\
    \\[-0.8em]
    Example 2 & \makecell[l]{
    Example 2:\\
    Text: An employee descending a scaffold missed the last rung and twisted his ankle. He was treated \\
    on site and returned to work the following day.\\
    Severity: non hospitalized injury}\\
    \\[-0.8em]
    ... & ...\\
    \\[-0.8em]
    Query & \makecell[l]{
    Question:\\
    Text: An employee was installing drywall when he cut his forearm on exposed metal framing. He was \\
    transported to a nearby hospital, where he remained overnight for treatment.}\\
    \\[-0.8em]
    Output & Severity: hospitalized injury
    \\[-0.8em]
    \\
    \bottomrule
\end{tabular}}
\caption{Example prompt used for Classification.}
\label{tab:prompt-cls}
\end{table*}

\end{document}